%% file: main.tex
\documentclass[conference]{IEEEtran}
\IEEEoverridecommandlockouts

\usepackage{cite}
\usepackage{amsmath,amssymb,amsfonts}
\usepackage{algorithmic}
\usepackage{graphicx}
\usepackage{textcomp}
\usepackage[dvipsnames]{xcolor}
\usepackage{lipsum}
\usepackage{amssymb}
\usepackage{tabularx} 
\usepackage{dblfloatfix}
\usepackage{placeins}
\usepackage{comment}
\usepackage[linesnumbered,ruled,vlined]{algorithm2e}
\usepackage[hidelinks]{hyperref}
\usepackage{lastpage}
\usepackage{dsfont}   
\usepackage{float}
\usepackage{enumitem}
\usepackage{mathrsfs}
\usepackage{mathtools}
\usepackage{tikz}
\usepackage{ctable}
\usepackage{balance}
\usepackage{multirow}
\usepackage{bbm}
\usepackage[utf8]{inputenc}
\usepackage{soul}
\sethlcolor{yellow}
\usepackage{subcaption}
\usepackage[numbers,sort&compress]{natbib}
\usepackage{soul}
\usepackage{setspace}
\usetikzlibrary{fit,calc}
\usepackage{colortbl}
\definecolor{paleyellow}{rgb}{1.0, 1.0, 0.8}
\sethlcolor{paleyellow}
\usepackage{soul}

\begin{document}

\title{Enhancing Remote Sensing Vision-Language Models for Zero-Shot Scene Classification

\thanks{\hspace{-3.5mm} * The authors have contributed equally to this work.
\vspace{1mm}
\\Acknowledgments -- M.Z. and B.G. are funded by the Walloon region under grant No. 2010235 (ARIAC by DIGITALWALLONIA4.AI). T.G. is funded by MedReSyst part of the Walloon Region and EU-Wallonie 2021-2027 program.}
}

\author{\IEEEauthorblockN{Karim El Khoury*$^{1}$\hspace{6mm}Maxime Zanella*$^{1,2}$\hspace{6mm}Beno\^{i}t Gérin*$^{1}$\hspace{6mm}Tiffanie Godelaine*$^{1}$\\Beno\^{i}t Macq$^{1}$\hspace{3mm}Sa\"id Mahmoudi$^{2}$\hspace{3mm}Christophe De Vleeschouwer$^{1}$\hspace{3mm}Ismail Ben Ayed\textit{$^{3}$}}

\\

\IEEEauthorblockA{$^{1}$UCLouvain, Belgium \hspace{6mm} $^{2}$UMons, Belgium \hspace{6mm} $^{3}$\'{E}TS Montreal, Canada
}

}

\maketitle

\begin{abstract}
Vision-Language Models for remote sensing have shown promising uses thanks to their extensive pretraining. However, their conventional usage in zero-shot scene classification methods still involves dividing large images into patches and making independent predictions, i.e., inductive inference, thereby limiting their effectiveness by ignoring valuable contextual information. Our approach tackles this issue by utilizing initial predictions based on text prompting and patch affinity relationships from the image encoder to enhance zero-shot capabilities through transductive inference, all without the need for supervision and at a minor computational cost. Experiments on 10 remote sensing datasets with state-of-the-art Vision-Language Models demonstrate significant accuracy improvements over inductive zero-shot classification. Our source code is publicly available on Github: \href{https://github.com/elkhouryk/RS-TransCLIP}{https://github.com/elkhouryk/RS-TransCLIP}

\end{abstract}

\vspace{0.5mm}


\input{Styles/Texts/1_introduction}

\input{Styles/Texts/2_related_work}

\input{Styles/Texts/3_method}
\input{Styles/Texts/4_experiments}

\input{Styles/Texts/5_conclusion}

\bibliographystyle{ieeetr}
{\small  
\bibliography{refs}
}

\end{document}

%% file: Styles/Texts/1_introduction.tex
\section{Introduction}

Remote Sensing (RS) imagery has become an effective tool for monitoring the surface of the Earth. It has given rise to several applications, ranging from environmental monitoring ~\cite{chen2023land, yuan2020deep}, to precision agriculture~\cite{ maes2019perspectives,phang2023satellite}, as well as emergency disaster response ~\cite{xia2023deep,streamlined2024fast}. All of these tasks require precise and quick scene classification to extract useful insights from highly complex visual data.

Linking images with text descriptions has been an effective approach for learning granular visual representations~\cite{sariyildiz2020learning, joulin2016learning}. While this idea seemed powerful, pioneering works in the field of RS~\cite{abdullah2020textrs, rahhal2020deep} were limited by computational budgets and the quantity of available RS data, both of which have been significant bottlenecks for generalization and robustness capabilities~\cite{liu2024remoteclip}. More recently, Vision-Language Models (VLMs) like CLIP \cite{pmlr-v139-radford21a} have overcome these limitations by leveraging a new pretraining paradigm that uses large-scale image-text pair datasets for unsupervised contrastive learning. These models have demonstrated high capability for numerous downstream tasks, including efficient zero-shot image classification by prompting arbitrary candidate class descriptions, e.g., \texttt{"a satellite photo of a [class]."}, sometimes even surpassing supervised competitors~\cite{pmlr-v139-radford21a}. Inspired by these promising results, the RS community has worked on developing large image-text RS datasets ~\cite{zhang2023rs5m,pang2024h2rsvlm,wang2024skyscript,Muhtar2024lhrs} leading to rapid progress in zero-shot scene classification benchmarks~\cite{10506064}.

\begin{figure}[t]
    \centering
    \includegraphics[width=0.95\linewidth]{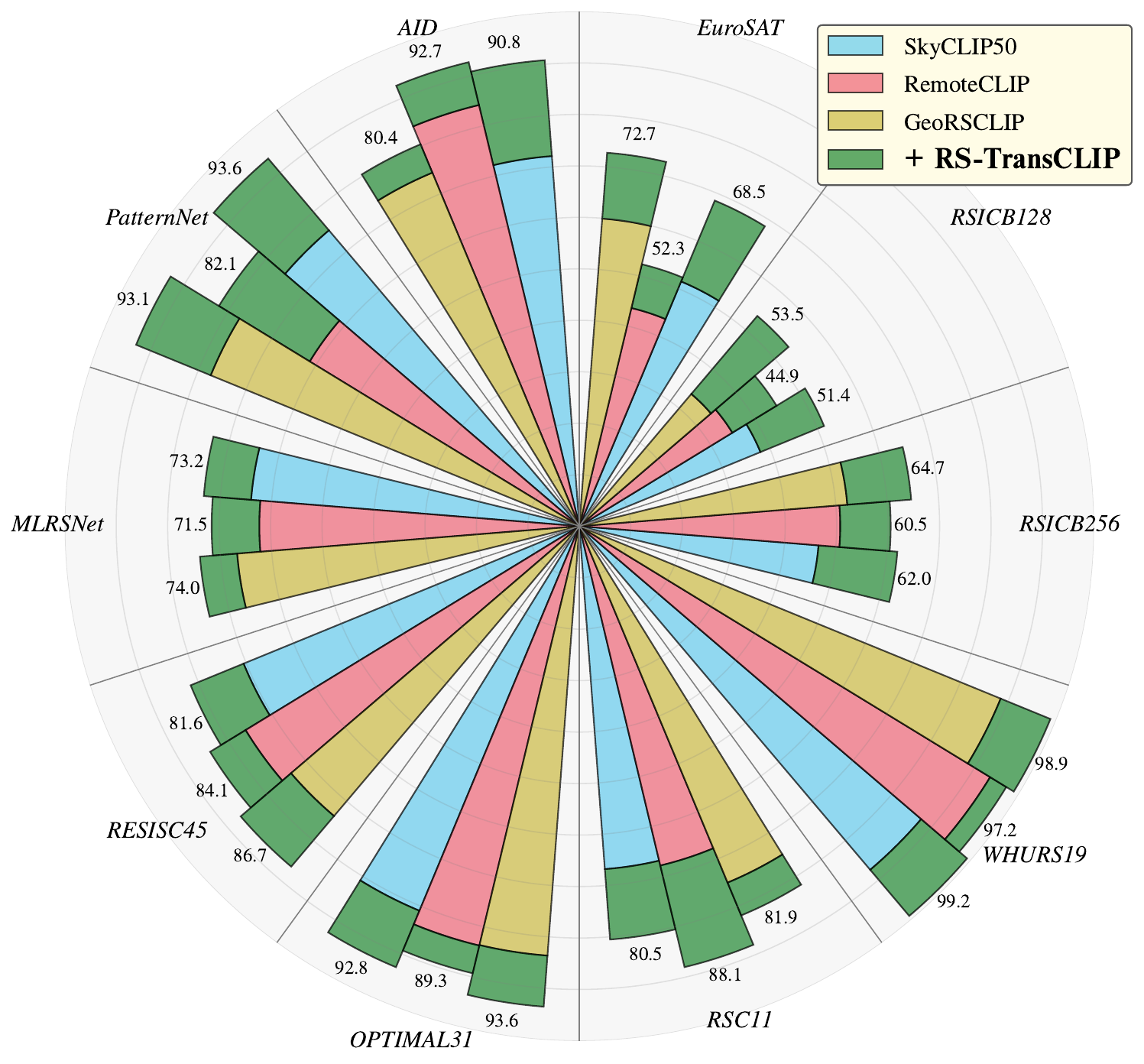}
    \caption{Top-1 accuracy of RS-TransCLIP, on ViT-L/14 RS VLMs, for zero-shot scene classification across 10 datasets.}
    \label{fig:polar}
\end{figure}



In remote sensing scene classification, both the large size of the images and the need for granular information pose challenges. To make high-resolution inference tractable, it is common practice to divide the images into smaller patches and generate predictions for each patch individually; this is known as \textit{inductive} inference. Another paradigm known as \textit{transductive} inference~\cite{788640, joachims1999transductive}, has shown that jointly considering multiple instances at prediction time can improve the prediction accuracy by accounting for the statistical distribution of instances in the embedding space~\cite{martin2024transductive, zanella2024boosting}. Despite its large potential, \textit{transductive} inference has been largely overlooked in RS within the context of VLMs. We aim to address this gap by introducing an efficient transductive method that operates exclusively within the embedding space, i.e., in a black box setup after feature extraction.

In a zero-shot classification setting, class-specific textual prompts are mapped to a shared embedding space generating individual pseudo-label for each image patch. In a traditional \textit{inductive} inference process, predictions are generated by utilizing initial pseudo-labels to identify the most confident class, with each patch predicted individually. In contrast, our work envisions \textit{transductive} inference in a zero-shot classification setting. As shown in Fig. \ref{fig:rs_transclip}, this approach leverages the data structure within the feature space to account for instance relations, enabling collective prediction of all points simultaneously. Our proposed objective function can be viewed as a regularized maximum-likelihood estimation, constrained by a Kullback-Leibler divergence penalty that integrates the aforementioned initial pseudo-labels and a Laplacian term that constraints similar patches to have similar predictions.



\textit{\textbf{Contribution:}} We introduce RS-TransCLIP, a transductive algorithm that enhances RS VLMs without requiring any labels, only incurring a negligible computational cost to the overall inference time. Fig. 1 highlights the significant boost that RS-TransCLIP offers on state-of-the-art RS VLMs.

%% file: Styles/Texts/2_related_work.tex
\medskip

\section{Related work}

\subsection{Vision-Language Models for Remote Sensing}

Due to foundation models being trained on natural images, there is an active research effort to build domain-specific versions of these models. This is prevalent in the medical imaging where VLMs have shown promising results~\cite{zhang2024biomed, eslami2021pubmedclip} in improving image-text retrieval and few-shot classification. The RS community has followed suit, working on creating extensive image-text datasets by scraping and filtering public satellite and UAV imagery sources~\cite{zhang2023rs5m, pang2024h2rsvlm,wang2024skyscript,Muhtar2024lhrs}. This has led to the development of several fine-tuned VLMs on various downstream tasks~\cite{luo2024skysensegpt,zhang2024earthgpt,mall2023graft,hu2023rsgpt,rs16091477}, with many of them showing strong performances in zero-shot scene classification ~\cite{liu2024remoteclip, zhang2023rs5m,wang2024skyscript}.

\subsection{Transductive inference in Vision-Language Models}

In the few-shot literature, transduction leverages both the few labeled samples and unlabeled test data outperforming inductive methods~\cite{dhillon2019baseline, boudiaf2020information, liu2020prototype, ziko2020laplacian}. However, when applied to VLMs, these transductive methods face significant performance drops~\cite{martin2024transductive, zanella2024boosting} since they are based solely on the vision features. This motivated very recent transductive methods in computer vision to explicitly leverage the textual modality alongside image embeddings -- a capability not present before the emergence of VLMs~\cite{martin2024transductive, zanella2024boosting, zanella2024histboosting}. Building on these advances and the transductive-inference zero-shot objective described in~\cite{zanella2024boosting}, our work enhances the predictive accuracy of pretrained RS VLMs without the need of any supervision.

%% file: Styles/Texts/3_method.tex
\vfill

\section{Method}
\label{sec:method}

\begin{figure}[t]
    \centering
    \includegraphics[width=\linewidth]{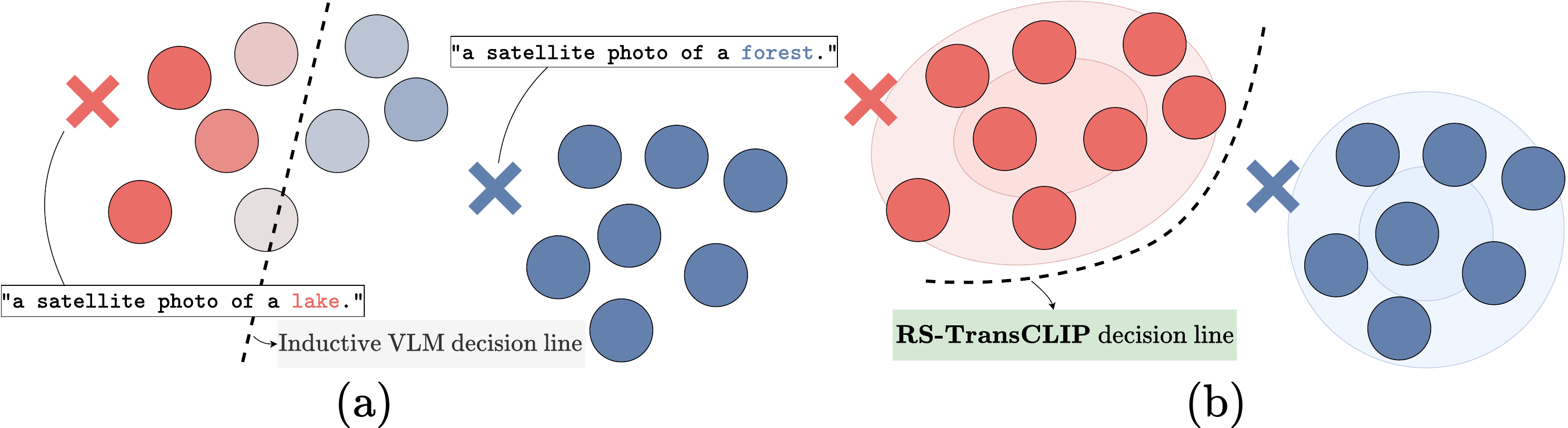}
    \caption{(a) VLMs assign each image to its closest text embedding and (b) RS-TransCLIP exploits the image-text structure to enhance the predictions without any additional labels.}
    \label{fig:rs_transclip}
\end{figure}

The transductive approach employed by RS-TransCLIP is based on the hypothesis that the data structure within the feature space can be modeled as a mixture of Gaussian distributions. As a result, the RS-TransCLIP objective function integrates this hypothesis alongside affinity relationships among patches and initial text-based pseudo-labels to minimize prediction deviation. The intuition behind the proposed transductive approach is depicted in Fig. \ref{fig:rs_transclip}.

\newpage

\subsection{Variable Definition}

In an \textit{inductive} approach, predictions are made individually using only the initial pseudo-label \(\mathbf{\hat{y}}\). Conversely, in the proposed \textit{transductive} approach, predictions are made simultaneously by modeling the feature space using three variables, \(\mathbf{z}\), \(\boldsymbol{\mu}\) and \(\boldsymbol{\Sigma}\) which can be split into two categories:\\


\noindent \textit{Assignment variables} ---  where \(\mathbf{z}\) is defined as:
\[
\mathbf{z}_i = (z_{i,k})_{1 \leq k \leq K} \in \Delta_K, \quad \forall i \in \mathcal{Q}
\]

with $K$ the number of classes, \(\mathcal{Q}\) the sample indices set and \(\Delta_K\) the K-dimensional probability simplex (prediction space).\\

\noindent \textit{Gaussian Mixture Model (GMM) variables} --- where the mean $\boldsymbol{\mu}$ and the covariance $\boldsymbol{\Sigma}$ are defined as: 
    \[
    \boldsymbol{\mu} = (\boldsymbol{\mu}_k \in \mathbb{R}^d)_{1 \leq k \leq K}
    \quad 
    \boldsymbol{\Sigma} = \operatorname{diag}(\sigma_1,\dots,\sigma_d)
    \]

 with $d$ the embedding dimension. Note that $\boldsymbol{\Sigma}$ is shared among classes to decrease the number of parameters.\\

\subsection{RS-TransCLIP objective function}

 The goal is to minimize the objective function ${\cal L}$ composed of three terms: an unsupervised \textit{GMM clustering} term, an affinity-based \textit{Laplacian regularization} term and a
divergence-driven \textit{Kullback-Leibler (KL) regularization} term. The terms of ${\cal L}$, written in Eq. \eqref{objective}, are detailed hereafter:

\begin{align}
\label{objective}
{\cal L} ({\mathbf z}, \boldsymbol{\mu}, \boldsymbol{\Sigma}) =  
& \underbrace{- \frac{1}{|{\cal Q}|} \sum_{i\in \cal Q}  {\mathbf z}_{i}^\top \log ({\mathbf p}_{i})}_{\textit{GMM clustering}} \\
& \underbrace{-\sum_{i\in \cal Q} \sum_{j \in \cal Q} w_{ij} \mathbf{z}_{i}^\top \mathbf{z}_{j}}_{\textit{Laplacian regularization}} \notag
& + \underbrace{\sum_{i \in \cal Q} \mbox{KL} (\mathbf{z}_{i} || \hat{\mathbf{y}}_{i})}_{\textit{KL regularization}} \notag
\end{align}

\vfill

\noindent \textit{GMM clustering} --- The goal of this term is maximizing the similarity between the assignment variables ${\mathbf z}_{i}$ and the likelihood ${\mathbf p}_{i}$. In our case, we model the likelihood of target data as a balanced mixture of $K$ multivariate Gaussian distributions. Each distribution represents a class \(k\) with an associated mean vector $\boldsymbol{\mu}_k$ and a covariance matrix $\boldsymbol{\Sigma}$. Defining ${\mathbf f}_i \in \mathbb{R}^d$ as the image embedding of sample $i$,
we set \(p_{i,k}\) the probability that sample $i$ is generated by the Gaussian distribution of class $k$:

\newpage

\[p_{i,k} \propto \det(\boldsymbol{\Sigma})^{-\frac{1}{2}} \exp \left(-{\frac {1}{2}}({\mathbf f}_i - \boldsymbol{\mu}_k)^\top \boldsymbol{\Sigma}^{-1}({\mathbf f}_i - \boldsymbol{\mu}_k) \right)\]

\medskip

\noindent \textit{Laplacian regularization} --- The aim of this term is to favor pairs of samples with high affinity to have similar assignment variables. In our case, we define non-negative affinities $w_{ij}$ using the cosine similarities between image embeddings of each sample (see line \ref{w} in Algorithm \ref{alg:RS_transclip}). Note that affinities can be tailored for each specific use-case, provided the affinity matrix is positive semi-definite. This ensures the concavity of the term, which in turn guarantees the convergence of the decoupled updates (refer to~\cite{zanella2024boosting} for details). \\

\noindent \textit{KL regularization} --- The purpose of this term is to prevent the assignment variables to deviate significantly from the initial pseudo-labels.  In our case, we obtain the pseudo-labels $(\hat{\mathbf{y}}_{i})_{1 \leq i \leq \mathcal{Q}}$ by applying the softmax function to the vector whose components are obtained by computing the dot product between the image embeddings $\mathbf{f}_i$ and all text embeddings $\mathbf{t}_k \in \mathbb{R}^d$, scaled by the temperature factor $\tau$ used during VLM pretraining (see line \ref{yhat} in Algorithm \ref{alg:RS_transclip}). This allows us to integrate the text-knowledge into the optimization process.

\subsection{Solving procedure}

We refer to~\cite{zanella2024boosting} for the derivation and optimization details of the convergence procedure for the objective function ${\cal L}$. The pseudo-code for the RS-TransCLIP procedure is outlined in Algorithm \ref{alg:RS_transclip}. Note that image and text embeddings are only computed once at the start. After the affinity $w_{ij}$ and the pseudo-labels $\hat{\mathbf{y}}_i$ are determined, the assignment variables $\mathbf{z}_{i}$ and the GMM variables ${\boldsymbol{\mu}}_k$ and ${\mathbf \Sigma}$ are then initialized and updated, according to the update rules listed in  Eq. \eqref{z-updates}, \eqref{u-update} and \eqref{sigma-update} respectively. The update rules vary depending on the two variable categories:\\

\noindent \textit{Iterative decoupled updates.} --- The assignment variable $\mathbf{z}_{i}$ is updated at each iteration $l$ as it depends on its neighbors $\mathbf{z}_{j}$. Note that the update rule of $\mathbf{z}_{i}$ can be parallelized, which makes the convergence procedure computationally efficient.

\begin{align}
\label{z-updates}
\mathbf{z}_{i}^{(l+1)} = \frac{\hat{\mathbf{y}}_{i} \odot \exp (\log (\mathbf{p}_{i}) + \sum_{j \in {\cal Q}} w_{ij} \mathbf{z}_{j}^{(l)})}{(\hat{\mathbf{y}}_{i} \odot \exp (\log (\mathbf{p}_{i}) + \sum_{j\in {\cal Q}} w_{ij}\mathbf{z}_{j}^{(l)}))^\top\mathbbm{1}_K}
\end{align}\\

\noindent \textit{Closed-form updates.} --- With 
$\mathbf{z}_{i}$ fixed, obtained following the iterative decoupled updates, we can calculate the closed-form updates for GMM variables ${\boldsymbol{\mu}}_k$ and ${\mathbf \Sigma}$.


\begin{equation}
\label{u-update}
    \boldsymbol{\mu}_k = \frac{\sum_{i \in {\cal Q}} z_{i,k}  {\mathbf f}_i}{ \sum_{i \in \cal Q} z_{i,k}}
\end{equation}

\begin{equation}
\label{sigma-update}
    \text{diag}({\mathbf \Sigma}) = \frac{\sum_{i \in \cal Q}\sum_{k} z_{i,k} (\mathbf{f}_i - \boldsymbol{\mu}_k)^2}{|\cal Q|}
\end{equation}

\definecolor{lightlime}{rgb}{0.9, 1.0, 0.9} 
\sethlcolor{lightlime}

\begin{algorithm}[!t]
\caption{RS-TransCLIP procedure}
\label{alg:RS_transclip}
\SetAlgoLined
\SetKwComment{Comment}{$\triangleright$\ }{}
\KwIn{$\mathbf{f}$, $\mathbf{t}$, $\mathbf{\tau}$}

$\hat{\mathbf{y}}_i \gets \textit{softmax}(\tau \mathbf{f}_i^\top \mathbf{t}) \quad \forall i$\label{yhat}\;

$w_{ij} = \mathbf{f}_i^\top \mathbf{f}_j \quad \forall i,j$\label{w}\;

$\mathbf{z}_i \gets \hat{\mathbf{y}}_i \quad \forall i$\;

Initialize $\boldsymbol{\mu}_k \quad \forall k, \text{and} \quad \text{diag}(\boldsymbol{\Sigma})$ \Comment*{See *}\
\While{not converged}{
    \tcp{\textcolor{Green}{Iterative decoupled updates}}
    \For{$l = 1:\dots$}{
        Update $\mathbf{z}^{(l+1)}_i \quad \forall i$ \Comment*{See Eq. \eqref{z-updates}}
    }
    \tcp{\textcolor{Green}{Closed-form updates}}
    Update $\boldsymbol{\mu}_k \quad \forall k$ \Comment*{See Eq. \eqref{u-update}}
    Update $\text{diag}(\boldsymbol{\Sigma})$ \Comment*{See Eq. \eqref{sigma-update}}
}
\textbf{return $\mathbf{z}$}\
\rule{0.915\linewidth}{0.4pt}
\footnotesize
* $\boldsymbol{\mu}_k$ is initialized by averaging the image embeddings of the 8 most confident samples according to the pseudo-labels, while $\text{diag}(\boldsymbol{\Sigma})$ is initialized by setting each element to $1/d$.

\end{algorithm}

%% file: Styles/Texts/4_experiments.tex
\section{Experiments}

\subsection{Experimental setup}

We test RS-TransCLIP on four VLMs: CLIP~\cite{pmlr-v139-radford21a}, RemoteCLIP~\cite{liu2024remoteclip}, SkyCLIP~\cite{wang2024skyscript}, and GeoRSCLIP~\cite{zhang2023rs5m} --- all with various model architectures to generate their respective image embeddings. Using RS text-prompt templates from~\cite{zhang2023rs5m}, 106 individual text embeddings were averaged out to get a single textual embedding per class. The zero-shot scene classification performance is evaluated on 10 RS benchmark datasets: AID, EuroSAT, MLRSNet, OPTIMAL31, PatternNet, RESISC45, RSC11, RSICB128, RSICB256, and WHURS19~\cite{xia2017aid,helber2018eurosat,qi2020mlrsnet,wang2019optimal,zhou2018patternnet,cheng2017resisc45,zhao2016rsc11,li2020rsicb,xia2010whurs19}. Note that none of the chosen VLMs were fine-tuned on any of the listed datasets. TABLE \ref{tab:model_performance} presents the zero-shot top-1 accuracy, \textit{without} and \textit{with} the addition of RS-TransCLIP.

\subsection{Zero-shot classification --- without RS-TransCLIP}

First, we assess the top-1 accuracy \textit{without} RS-TransCLIP, evaluating it in an \textit{inductive} inference scenario based on the initial pseudo-labels $\hat{\mathbf{y}}_i$ (see line 1 in Algorithm 1). We notice that for smaller backbones like ViT-B/32, RemoteCLIP, GeoRSCLIP and SkyCLIP50 outperform CLIP. However, for larger backbones like ViT-L/14, CLIP is surprisingly competitive on various benchmarks in comparison to the RS VLMs. A clear trend of larger models performing better indicates promising potential in scaling both model and dataset sizes.

\input{Styles/Tables/ZS_RS-TransCLIP_averageprompt}

\subsection{Zero-shot classification — with RS-TransCLIP}

Second, we observe the top-1 accuracy \textit{with} RS-TransCLIP, evaluating it in a \textit{transductive} inference scenario based on the obtained assignment variables $\mathbf{z}_i$ when solving ${\cal L}$ (see Algorithm 1). We can clearly see a massive performance improvement across all benchmarks and models. We find that the addition of RS-TransCLIP provides average gains ranging from 9.9\% up to 17.1\% across all benchmarks and models.

\newpage

Interestingly, RS-TransCLIP produces notable improvements even when the inductive model's top-1 accuracy performance is already high. For example, when GeoRSCLIP ViT-H/14 is applied to WHURS19, the top-1 accuracy increases from 90.4\% to 99.7\%. Similarly, for the same model applied to PatternNet, the top-1 accuracy improves from 82.7\% to 96.2\%. This shows RS-TransCLIP's applicability for tasks where these VLMs are already effective, without any labels.

We also notice that, for the ViT-L/14 backbone, RS-TransCLIP offers slightly higher gains to SkyCLIP50 compared to CLIP and RemoteCLIP, allowing it to outperform them when combined with transduction. RS-TransCLIP also demonstrates its applicability to more robust models, bringing an average gain of 12.1\% on GeoRSCLIP ViT-H/14.

\vfill

\subsection{RS-TransCLIP computational cost}

We evaluated the computational cost of RS-TransCLIP using three datasets of varying sizes. As shown in TABLE \ref{tab:comp_cost}, the feature extraction time increases with the number of image patches while the additional load from RS-TransCLIP remains minimal. Thus, by not requiring optimization of model parameters or input prompts \cite{Zanella_2024_CVPR}, our transductive method ensures fast inference all while boosting model accuracy.

\newpage

\input{Styles/Tables/comp_cost_RS-TransCLIP}

%% file: Styles/Tables/ZS_RS-TransCLIP_averageprompt.tex
\definecolor{green}{rgb}{0, 0.6, 0.2} 
\definecolor{red}{rgb}{0.7, 0, 0} 
\definecolor{blue}{rgb}{0.93, 0.97, 0.985} 

\begin{table*}[!ht]

    \caption{Top-1 accuracy for zero-shot scene classification without (white) and with (blue) RS-TransCLIP on 10 RS datasets.}

    \centering
    \resizebox{\textwidth}{!}{\begin{tabular}{llcccccccccccc}
        \toprule
         & \textbf{Model} & \textbf{AID} & \textbf{EuroSAT} & \textbf{MLRSNet} & \textbf{OPTIMAL31} & \textbf{PatternNet} & \textbf{RESISC45} & \textbf{RSC11} & \textbf{RSICB128} & \textbf{RSICB256} & \textbf{WHURS19} & \textbf{Average} \\
        \midrule
        \parbox[t]{2mm}{\multirow{6}{*}{\rotatebox[origin=c]{90}{\textbf{ResNet-50}}}} & CLIP & 55.4 & 28.3 & 45.0 & 64.5 & 46.4 & 52.8 & 56.7 & 23.4 & 30.4 & 71.3 & 47.4 \\
         & \cellcolor{blue}\textbf{+ RS-TransCLIP} & \cellcolor{blue}\textbf{69.6} & \cellcolor{blue}\textbf{48.1} & \cellcolor{blue}\textbf{54.2} & \cellcolor{blue}\textbf{79.6} & \cellcolor{blue}\textbf{69.0} & \cellcolor{blue}\textbf{69.6} & \cellcolor{blue}\textbf{77.8} & \cellcolor{blue}\textbf{34.3} & \cellcolor{blue}\textbf{46.8} & \cellcolor{blue}\textbf{95.9} & \cellcolor{blue}\textbf{64.5} \\
         & \cellcolor{blue}$\Delta$ & \cellcolor{blue}\textcolor{green}{\textbf{+14.2}} & \cellcolor{blue}\textcolor{green}{\textbf{+19.8}} & \cellcolor{blue}\textcolor{green}{\textbf{+9.3}} & \cellcolor{blue}\textcolor{green}{\textbf{+15.2}} & \cellcolor{blue}\textcolor{green}{\textbf{+22.6}} & \cellcolor{blue}\textcolor{green}{\textbf{+16.7}} & \cellcolor{blue}\textcolor{green}{\textbf{+21.0}} & \cellcolor{blue}\textcolor{green}{\textbf{+10.8}} & \cellcolor{blue}\textcolor{green}{\textbf{+16.4}} & \cellcolor{blue}\textcolor{green}{\textbf{+24.6}} & \cellcolor{blue}\textcolor{green}{\textbf{+17.1}} \\
         \cmidrule{2-13}
         & RemoteCLIP & 89.1 & 26.7 & 43.0 & 64.0 & 43.6 & 51.6 & 67.0 & 15.0 & 36.4 & 95.4 & 53.2 \\
         & \cellcolor{blue}\textbf{+ RS-TransCLIP} & \cellcolor{blue}\textbf{93.3} & \cellcolor{blue}\textbf{34.4} & \cellcolor{blue}\textbf{58.0} & \cellcolor{blue}\textbf{85.0} & \cellcolor{blue}\textbf{53.6} & \cellcolor{blue}\textbf{72.9} & \cellcolor{blue}\textbf{87.2} & \cellcolor{blue}\textbf{19.1} & \cellcolor{blue}\textbf{48.2} & \cellcolor{blue}\textbf{98.4} & \cellcolor{blue}\textbf{65.0} \\
         & \cellcolor{blue}$\Delta$ & \cellcolor{blue}\textcolor{green}{\textbf{+4.2}} & \cellcolor{blue}\textcolor{green}{\textbf{+7.8}} & \cellcolor{blue}\textcolor{green}{\textbf{+15.0}} & \cellcolor{blue}\textcolor{green}{\textbf{+21.0}} & \cellcolor{blue}\textcolor{green}{\textbf{+10.0}} & \cellcolor{blue}\textcolor{green}{\textbf{+21.2}} & \cellcolor{blue}\textcolor{green}{\textbf{+20.2}} & \cellcolor{blue}\textcolor{green}{\textbf{+4.1}} & \cellcolor{blue}\textcolor{green}{\textbf{+11.8}} & \cellcolor{blue}\textcolor{green}{\textbf{+3.0}} & \cellcolor{blue}\textcolor{green}{\textbf{+11.8}} \\
        \midrule\midrule
        \parbox[t]{2mm}{\multirow{12}{*}{\rotatebox[origin=c]{90}{\textbf{ViT-B/32}}}} & CLIP & 66.4 & 45.3 & 51.2 & 73.0 & 59.6 & 60.7 & 55.5 & 27.7 & 40.3 & 81.1 & 56.1 \\
         & \cellcolor{blue}\textbf{+ RS-TransCLIP} & \cellcolor{blue}\textbf{80.7} & \cellcolor{blue}\textbf{49.0} & \cellcolor{blue}\textbf{64.2} & \cellcolor{blue}\textbf{82.9} & \cellcolor{blue}\textbf{76.6} & \cellcolor{blue}\textbf{74.1} & \cellcolor{blue}\textbf{67.0} & \cellcolor{blue}\textbf{33.2} & \cellcolor{blue}\textbf{46.4} & \cellcolor{blue}\textbf{90.3} & \cellcolor{blue}\textbf{66.5} \\
         & \cellcolor{blue}$\Delta$ & \cellcolor{blue}\textcolor{green}{\textbf{+14.3}} & \cellcolor{blue}\textcolor{green}{\textbf{+3.6}} & \cellcolor{blue}\textcolor{green}{\textbf{+13.0}} & \cellcolor{blue}\textcolor{green}{\textbf{+9.9}} & \cellcolor{blue}\textcolor{green}{\textbf{+16.9}} & \cellcolor{blue}\textcolor{green}{\textbf{+13.4}} & \cellcolor{blue}\textcolor{green}{\textbf{+11.5}} & \cellcolor{blue}\textcolor{green}{\textbf{+5.6}} & \cellcolor{blue}\textcolor{green}{\textbf{+6.0}} & \cellcolor{blue}\textcolor{green}{\textbf{+9.3}} & \cellcolor{blue}\textcolor{green}{\textbf{+10.4}} \\
        \cmidrule{2-13}
         & GeoRSCLIP & 70.3 & 53.4 & 65.0 & 79.6 & 75.8 & 68.8 & 68.3 & 29.0 & 46.5 & 88.8 & 64.5 \\
         & \cellcolor{blue}\textbf{+ RS-TransCLIP} & \cellcolor{blue}\textbf{78.2} & \cellcolor{blue}\textbf{69.0} & \cellcolor{blue}\textbf{71.9} & \cellcolor{blue}\textbf{87.3} & \cellcolor{blue}\textbf{94.5} & \cellcolor{blue}\textbf{79.5} & \cellcolor{blue}\textbf{78.6} & \cellcolor{blue}\textbf{42.8} & \cellcolor{blue}\textbf{61.8} & \cellcolor{blue}\textbf{98.7} & \cellcolor{blue}\textbf{76.2} \\
         & \cellcolor{blue}$\Delta$ & \cellcolor{blue}\textcolor{green}{\textbf{+7.9}} & \cellcolor{blue}\textcolor{green}{\textbf{+15.5}} & \cellcolor{blue}\textcolor{green}{\textbf{+6.9}} & \cellcolor{blue}\textcolor{green}{\textbf{+7.7}} & \cellcolor{blue}\textcolor{green}{\textbf{+18.6}} & \cellcolor{blue}\textcolor{green}{\textbf{+10.7}} & \cellcolor{blue}\textcolor{green}{\textbf{+10.3}} & \cellcolor{blue}\textcolor{green}{\textbf{+13.8}} & \cellcolor{blue}\textcolor{green}{\textbf{+15.3}} & \cellcolor{blue}\textcolor{green}{\textbf{+10.0}} & \cellcolor{blue}\textcolor{green}{\textbf{+11.7}} \\
        \cmidrule{2-13}
         & RemoteCLIP & 91.7 & 35.5 & 56.3 & 77.6 & 55.9 & 68.1 & 61.8 & 26.0 & 41.5 & 95.2 & 61.0 \\
         & \cellcolor{blue}\textbf{+ RS-TransCLIP} & \cellcolor{blue}\textbf{95.6} & \cellcolor{blue}\textbf{51.0} & \cellcolor{blue}\textbf{65.8} & \cellcolor{blue}\textbf{87.8} & \cellcolor{blue}\textbf{70.7} & \cellcolor{blue}\textbf{79.4} & \cellcolor{blue}\textbf{79.7} & \cellcolor{blue}\textbf{31.1} & \cellcolor{blue}\textbf{49.2} & \cellcolor{blue}\textbf{97.9} & \cellcolor{blue}\textbf{70.8} \\
         & \cellcolor{blue}$\Delta$ & \cellcolor{blue}\textcolor{green}{\textbf{+3.9}} & \cellcolor{blue}\textcolor{green}{\textbf{+15.5}} & \cellcolor{blue}\textcolor{green}{\textbf{+9.5}} & \cellcolor{blue}\textcolor{green}{\textbf{+10.3}} & \cellcolor{blue}\textcolor{green}{\textbf{+14.8}} & \cellcolor{blue}\textcolor{green}{\textbf{+11.2}} & \cellcolor{blue}\textcolor{green}{\textbf{+17.9}} & \cellcolor{blue}\textcolor{green}{\textbf{+5.1}} & \cellcolor{blue}\textcolor{green}{\textbf{+7.7}} & \cellcolor{blue}\textcolor{green}{\textbf{+2.7}} & \cellcolor{blue}\textcolor{green}{\textbf{+9.9}} \\
        \cmidrule{2-13}
         & SkyCLIP50 & 70.3 & 52.6 & 63.2 & 79.5 & 73.8 & 66.7 & 61.2 & 39.0 & 47.1 & 91.0 & 64.5 \\
         & \cellcolor{blue}\textbf{+ RS-TransCLIP} & \cellcolor{blue}\textbf{78.7} & \cellcolor{blue}\textbf{64.5} & \cellcolor{blue}\textbf{73.2} & \cellcolor{blue}\textbf{85.2} & \cellcolor{blue}\textbf{87.6} & \cellcolor{blue}\textbf{77.3} & \cellcolor{blue}\textbf{77.1} & \cellcolor{blue}\textbf{49.4} & \cellcolor{blue}\textbf{59.1} & \cellcolor{blue}\textbf{97.8} & \cellcolor{blue}\textbf{75.0} \\
         & \cellcolor{blue}$\Delta$ & \cellcolor{blue}\textcolor{green}{\textbf{+8.3}} & \cellcolor{blue}\textcolor{green}{\textbf{+11.9}} & \cellcolor{blue}\textcolor{green}{\textbf{+10.1}} & \cellcolor{blue}\textcolor{green}{\textbf{+5.8}} & \cellcolor{blue}\textcolor{green}{\textbf{+13.8}} & \cellcolor{blue}\textcolor{green}{\textbf{+10.6}} & \cellcolor{blue}\textcolor{green}{\textbf{+15.9}} & \cellcolor{blue}\textcolor{green}{\textbf{+10.4}} & \cellcolor{blue}\textcolor{green}{\textbf{+11.9}} & \cellcolor{blue}\textcolor{green}{\textbf{+6.8}} & \cellcolor{blue}\textcolor{green}{\textbf{+10.5}} \\
        \midrule\midrule

        \parbox[t]{2mm}{\multirow{12}{*}{\rotatebox[origin=c]{90}{\textbf{ViT-L/14}}}} & CLIP & 69.7 & 60.1 & 64.1 & 80.6 & 74.7 & 71.3 & 67.3 & 37.9 & 47.2 & 85.5 & 65.8 \\
        & \cellcolor{blue}\textbf{+ RS-TransCLIP} & \cellcolor{blue}\textbf{84.2} & \cellcolor{blue}\textbf{71.9} & \cellcolor{blue}\textbf{74.5} & \cellcolor{blue}\textbf{92.4} & \cellcolor{blue}\textbf{91.8} & \cellcolor{blue}\textbf{82.2} & \cellcolor{blue}\textbf{80.5} & \cellcolor{blue}\textbf{43.9} & \cellcolor{blue}\textbf{50.5} & \cellcolor{blue}\textbf{99.1} & \cellcolor{blue}\textbf{77.1} \\
         & \cellcolor{blue}$\Delta$ & \cellcolor{blue}\textcolor{green}{\textbf{+14.4}} & \cellcolor{blue}\textcolor{green}{\textbf{+11.9}} & \cellcolor{blue}\textcolor{green}{\textbf{+10.4}} & \cellcolor{blue}\textcolor{green}{\textbf{+11.7}} & \cellcolor{blue}\textcolor{green}{\textbf{+17.1}} & \cellcolor{blue}\textcolor{green}{\textbf{+10.9}} & \cellcolor{blue}\textcolor{green}{\textbf{+13.2}} & \cellcolor{blue}\textcolor{green}{\textbf{+5.9}} & \cellcolor{blue}\textcolor{green}{\textbf{+3.3}} & \cellcolor{blue}\textcolor{green}{\color{green}{\textbf{+13.6}}} & \cellcolor{blue}\textcolor{green}{\color{green}{\textbf{+11.3}}} \\
        \cmidrule{2-13}
         & GeoRSCLIP & 74.4 & 59.9 & 66.7 & 83.7 & 77.4 & 73.8 & 75.0 & 33.7 & 52.2 & 88.5 & 68.5 \\
         & \cellcolor{blue}\textbf{+ RS-TransCLIP} & \cellcolor{blue}\textbf{80.4} & \cellcolor{blue}\textbf{72.7} & \cellcolor{blue}\textbf{74.0} & \cellcolor{blue}\textbf{93.6} & \cellcolor{blue}\textbf{93.1} & \cellcolor{blue}\textbf{86.7} & \cellcolor{blue}\textbf{81.9} & \cellcolor{blue}\textbf{53.5} & \cellcolor{blue}\textbf{64.7} & \cellcolor{blue}\textbf{98.9} & \cellcolor{blue}\textbf{79.9} \\
         & \cellcolor{blue}$\Delta$ & \cellcolor{blue}\textcolor{green}{\textbf{+6.0}} & \cellcolor{blue}\textcolor{green}{\textbf{+12.8}} & \cellcolor{blue}\textcolor{green}{\textbf{+7.3}} & \cellcolor{blue}\textcolor{green}{\textbf{+9.9}} & \cellcolor{blue}\textcolor{green}{\textbf{+15.7}} & \cellcolor{blue}\textcolor{green}{\textbf{+12.9}} & \cellcolor{blue}\textcolor{green}{\textbf{+6.9}} & \cellcolor{blue}\textcolor{green}{\textbf{+19.9}} & \cellcolor{blue}\textcolor{green}{\textbf{+12.4}} & \cellcolor{blue}\textcolor{green}{\textbf{+10.4}} & \cellcolor{blue}\textcolor{green}{\textbf{+11.4}} \\
        \cmidrule{2-13}
         & RemoteCLIP & 84.1 & 43.6 & 62.2 & 83.8 & 61.4 & 76.0 & 67.8 & 34.8 & 50.7 & 93.5 & 65.8 \\
        & \cellcolor{blue}\textbf{+ RS-TransCLIP} & \cellcolor{blue}\textbf{92.7} & \cellcolor{blue}\textbf{52.3} & \cellcolor{blue}\textbf{71.5} & \cellcolor{blue}\textbf{89.3} & \cellcolor{blue}\textbf{82.1} & \cellcolor{blue}\textbf{84.1} & \cellcolor{blue}\textbf{88.1} & \cellcolor{blue}\textbf{44.9} & \cellcolor{blue}\textbf{60.5} & \cellcolor{blue}\textbf{97.2} & \cellcolor{blue}\textbf{76.3} \\
         & \cellcolor{blue}$\Delta$ & \cellcolor{blue}\textcolor{green}{\textbf{+8.6}} & \cellcolor{blue}\textcolor{green}{\textbf{+8.7}} & \cellcolor{blue}\textcolor{green}{\textbf{+9.3}} & \cellcolor{blue}\textcolor{green}{\textbf{+5.5}} & \cellcolor{blue}\textcolor{green}{\textbf{+20.7}} & \cellcolor{blue}\textcolor{green}{\textbf{+8.1}} & \cellcolor{blue}\textcolor{green}{\textbf{+20.3}} & \cellcolor{blue}\textcolor{green}{\textbf{+10.1}} & \cellcolor{blue}\textcolor{green}{\textbf{+9.8}} & \cellcolor{blue}\textcolor{green}{\textbf{+3.7}} & \cellcolor{blue}\textcolor{green}{\textbf{+10.5}} \\
        \cmidrule{2-13}
         & SkyCLIP50 & 72.1 & 51.5 & 64.0 & 80.9 & 75.3 & 70.5 & 66.8 & 38.0 & 46.6 & 87.5 & 65.3 \\
         & \cellcolor{blue}\textbf{+ RS-TransCLIP} & \cellcolor{blue}\textbf{90.8} & \cellcolor{blue}\textbf{68.5} & \cellcolor{blue}\textbf{73.2} & \cellcolor{blue}\textbf{92.8} & \cellcolor{blue}\textbf{93.6} & \cellcolor{blue}\textbf{81.6} & \cellcolor{blue}\textbf{80.5} & \cellcolor{blue}\textbf{51.4} & \cellcolor{blue}\textbf{62.0} & \cellcolor{blue}\textbf{99.2} & \cellcolor{blue}\textbf{79.4} \\
         & \cellcolor{blue}$\Delta$ & \cellcolor{blue}\textcolor{green}{\textbf{+18.7}} & \cellcolor{blue}\textcolor{green}{\textbf{+17.0}} & \cellcolor{blue}\textcolor{green}{\textbf{+9.2}} & \cellcolor{blue}\textcolor{green}{\textbf{+11.9}} & \cellcolor{blue}\textcolor{green}{\textbf{+18.3}} & \cellcolor{blue}\textcolor{green}{\textbf{+11.1}} & \cellcolor{blue}\textcolor{green}{\textbf{+13.7}} & \cellcolor{blue}\textcolor{green}{\textbf{+13.4}} & \cellcolor{blue}\textcolor{green}{\textbf{+15.4}} & \cellcolor{blue}\textcolor{green}{\textbf{+11.7}} & \cellcolor{blue}\textcolor{green}{\textbf{+14.0}} \\
        \midrule\midrule     
        \parbox[t]{2mm}{\multirow{3}{*}{\rotatebox[origin=c]{90}{\textbf{ViT-H/14}\hspace{-0.6mm}}}} & GeoRSCLIP & 76.3 & 68.3 & 67.4 & 84.8 & 82.7 & 73.8 & 77.4 & 43.1 & 56.5 & 90.4 & 72.1 \\
         & \cellcolor{blue}\textbf{+ RS-TransCLIP} & \cellcolor{blue}\textbf{83.8} & \cellcolor{blue}\textbf{91.2} & \cellcolor{blue}\textbf{78.1} & \cellcolor{blue}\textbf{94.5} & \cellcolor{blue}\textbf{96.2} & \cellcolor{blue}\textbf{88.0} & \cellcolor{blue}\textbf{83.3} & \cellcolor{blue}\textbf{54.8} & \cellcolor{blue}\textbf{72.8} & \cellcolor{blue}\textbf{99.7} & \cellcolor{blue}\textbf{84.2} \\
         & \cellcolor{blue}$\Delta$ & \cellcolor{blue}\textcolor{green}{\textbf{+7.5}} & \cellcolor{blue}\textcolor{green}{\textbf{+22.9}} & \cellcolor{blue}\textcolor{green}{\textbf{+10.7}} & \cellcolor{blue}\textcolor{green}{\textbf{+9.7}} & \cellcolor{blue}\textcolor{green}{\textbf{+13.5}} & \cellcolor{blue}\textcolor{green}{\textbf{+14.2}} & \cellcolor{blue}\textcolor{green}{\textbf{+5.9}} & \cellcolor{blue}\textcolor{green}{\textbf{+11.7}} & \cellcolor{blue}\textcolor{green}{\textbf{+16.3}} & \cellcolor{blue}\textcolor{green}{\textbf{+9.3}} & \cellcolor{blue}\textcolor{green}{\textbf{+12.1}} \\
        \bottomrule
    \end{tabular}}
    \label{tab:model_performance}

\end{table*}

%% file: Styles/Tables/comp_cost_RS-TransCLIP.tex
\begin{table}[t]
    \centering
    \footnotesize
    \vspace{1.2mm}
    \caption{RS-TransCLIP run time on top of CLIP ViT-L/14, evaluated with 24GB NVIDIA GeForce RTX 4090 GPU.}

    \resizebox{\linewidth}{!}{\begin{tabular}{cccc}
        \toprule
        \textbf{RS} & \textbf{Total} & \textbf{Features encoding} & \textbf{+ RS-TransCLIP}\\
        \textbf{dataset} & \textbf{patches} & \textbf{time} & \textbf{time} \\
        \midrule
        WHURS19 & $\thicksim10^3$ & $\thicksim \text{8 seconds}$ & $\thicksim \text{0.3 seconds}$\\

        AID & $\thicksim10^4$ & $\thicksim \text{40 seconds}$ & $\thicksim \text{2 seconds}$\\
    
        MLRSNet & $\thicksim10^5$ & $\thicksim \text{6 minutes}$ & $\thicksim \text{25 seconds}$\\
    
        \bottomrule
    \end{tabular}}
    \label{tab:comp_cost}

\end{table}

%% file: Styles/Texts/5_conclusion.tex
\vfill

\section{Conclusion}

In this work, we proposed RS-TransCLIP, a transductive algorithm that enhances RS VLMs with minimal extra computational cost. By leveraging initial pseudo-labels and patch affinities, our method improves zero-shot capabilities through \textit{transductive} inference, demonstrating significant accuracy improvements over inductive zero-shot classification and showing its wide applicability beyond natural images \cite{zanella2024boosting}. Future works will study RS-TransCLIP's performance concerning text-prompt variability, given VLMs' high sensitivity to input text prompts. Moreover, adapting RS-TransCLIP to a few-shot setting to incorporate labeled data will be explored in human-in-the-loop scenarios.